\documentclass[letterpaper]{article}
\usepackage[preprint]{aaai2027}
\usepackage[hyphens]{url}
\usepackage{graphicx}
\usepackage{natbib}
\usepackage{caption}
\usepackage{booktabs}
\usepackage{amsmath}
\usepackage{amssymb}
\usepackage{amsfonts}
\usepackage{mathtools}
\usepackage{makecell}
\usepackage{array}
\usepackage{multirow}
\title{Enhancing Irregular Time Series Forecasting with\\ Continuous-Time Modeling Framework}
\author{
Tianen Shen, Zhengyu Li, Yutong Li, Xiangfei Qiu,\\
Xingjian Wu, Bin Yang, Jilin Hu
}

\affiliations{}
\begin{document}
\maketitle
\begin{abstract}
Irregular multivariate time series are widely encountered in applications such as healthcare monitoring, human activity recognition, and environmental sensing. Their core challenges stem from asynchronous observations, non-uniform sampling intervals, and the fact that temporal patterns themselves carry critical dynamic information. Existing approaches either rely on discretization-based preprocessing (e.g., interpolation, imputation, or aggregation), which disrupts the underlying continuous-time semantics, or adopt continuous-time modeling via ODE-based frameworks, which typically require specialized architectures and incur substantial computational overhead due to numerical solvers. To address these limitations, we propose WrapFlow, a continuous-time modeling framework for irregular time series forecasting. On the input side, WrapFlow introduces Continuous-Time Tokenization, which directly encodes raw observation events and explicitly models long unobserved intervals via gap-aware tokens. The resulting continuous-time tokens are then processed by a standard Transformer backbone to capture long-range temporal dependencies. On the output side, we develop a simulation-free training paradigm for Residual Flow Matching, which learns conditional residual vector fields around base predictions while avoiding numerical-solver simulation and backpropagation during training. This design enables high-quality continuous forecasting using only a small number of fixed rollout steps at inference. Extensive experiments on multiple real-world datasets demonstrate that WrapFlow achieves state-of-the-art performance.

\end{abstract}
\section{Introduction}
\label{sec:Introduction}
Irregular multivariate time series (IMTS) are widely observed in applications such as healthcare monitoring, human activity recognition, and environmental sensing~\citep{Yao2018,de2019gru,wu2025k2vae,gao2025ssdts,liu2026astgi}. Unlike regularly sampled time series, IMTS consist of asynchronous observation events, where different variables are recorded at different timestamps with highly non-uniform intervals~\citep{Neural-CDE,liu2026rethinking,qiu2025duet}. In this setting, time is not merely an index for observations, but a critical signal that reflects system dynamics. For instance, dense observations may indicate rapid state changes, while long gaps may imply stability. \textit{Therefore, irregular multivariate time series forecasting (IMTSF) is more naturally formulated as a continuous-time modeling problem, where models must capture not only observed values but also the timing and intervals between observations.}

Existing approaches for IMTSF generally fall into two categories. The first relies on discretization-based preprocessing~\citep{Zhang2024TPatchGNN,MTAN,wang2025optimal,Luo2025HiPatch} (e.g., interpolation, imputation, or temporal aggregation) to align irregular data onto a fixed time grid, followed by standard forecasting methods. While simple and compatible, such methods fail to account for continuous-time modeling and often irreversibly distort the original temporal semantics~\citep{liu2026astgi}. The second category adopts continuous-time modeling, typically based on ordinary differential equations (ODEs)~\citep{schirmer2022modeling,bilovs2021neural}, to jointly model historical evolution and future prediction. Although more principled, these methods usually depend on specialized dynamic architectures and numerical solvers, resulting in high computational cost and cumbersome optimization.

The recognition of the above challenges naturally raises a pivotal question:

\begingroup
\makeatletter
\newsavebox{\ctflowquestionbox}
\newdimen\ctflowWdim
\newdimen\ctflowTdim
\newdimen\ctflowBdim
\newdimen\ctflowRdim
\newdimen\ctflowKdim
\newdimen\ctflowWRdim
\newdimen\ctflowWRKdim
\newdimen\ctflowTRdim
\newdimen\ctflowTRKdim
\newdimen\ctflowBRdim
\newdimen\ctflowBRKdim
\newdimen\ctflowRKdim
\setlength{\fboxsep}{5pt}
\sbox{\ctflowquestionbox}{%
  \parbox{\dimexpr\linewidth-2\fboxsep-4pt\relax}{%
    \centering\small\itshape
    How can IMTSF models capture continuous-time dynamics efficiently and effectively?%
  }%
}%
\ctflowWdim=\wd\ctflowquestionbox
\advance\ctflowWdim by 2\fboxsep
\ctflowTdim=\ht\ctflowquestionbox
\advance\ctflowTdim by \fboxsep
\ctflowBdim=-\dp\ctflowquestionbox
\advance\ctflowBdim by -\fboxsep
\ctflowRdim=6pt
\ctflowKdim=.55228475\ctflowRdim
\ctflowWRdim=\ctflowWdim \advance\ctflowWRdim by -\ctflowRdim
\ctflowWRKdim=\ctflowWRdim \advance\ctflowWRKdim by \ctflowKdim
\ctflowTRdim=\ctflowTdim \advance\ctflowTRdim by -\ctflowRdim
\ctflowTRKdim=\ctflowTRdim \advance\ctflowTRKdim by \ctflowKdim
\ctflowBRdim=\ctflowBdim \advance\ctflowBRdim by \ctflowRdim
\ctflowBRKdim=\ctflowBRdim \advance\ctflowBRKdim by -\ctflowKdim
\ctflowRKdim=\ctflowRdim \advance\ctflowRKdim by -\ctflowKdim
\edef\ctW{\strip@pt\ctflowWdim}
\edef\ctT{\strip@pt\ctflowTdim}
\edef\ctB{\strip@pt\ctflowBdim}
\edef\ctR{\strip@pt\ctflowRdim}
\edef\ctWR{\strip@pt\ctflowWRdim}
\edef\ctWRK{\strip@pt\ctflowWRKdim}
\edef\ctTR{\strip@pt\ctflowTRdim}
\edef\ctTRK{\strip@pt\ctflowTRKdim}
\edef\ctBR{\strip@pt\ctflowBRdim}
\edef\ctBRK{\strip@pt\ctflowBRKdim}
\edef\ctRK{\strip@pt\ctflowRKdim}
\begin{center}
\leavevmode
\hbox{%
  \vrule width0pt height\ctflowTdim depth-\ctflowBdim
  \pdfliteral {q 0.47 G 1.3 w
    \ctR\space \ctB\space m
    \ctWR\space \ctB\space l
    \ctWRK\space \ctB\space \ctW\space \ctBRK\space \ctW\space \ctBR\space c
    \ctW\space \ctTR\space l
    \ctW\space \ctTRK\space \ctWRK\space \ctT\space \ctWR\space \ctT\space c
    \ctR\space \ctT\space l
    \ctRK\space \ctT\space 0 \ctTRK\space 0 \ctTR\space c
    0 \ctBR\space l
    0 \ctBRK\space \ctRK\space \ctB\space \ctR\space \ctB\space c
    h S Q}%
  \kern\fboxsep
  \usebox{\ctflowquestionbox}%
  \kern\fboxsep
}%
\end{center}
\makeatother
\endgroup

To address this question, we propose \textit{WrapFlow}: a continuous-time modeling framework for irregular time series forecasting. WrapFlow consists of two complementary components: input-side representation modeling and output-side continuous generation, connected by a Transformer backbone for sequence modeling. On the input side, we introduce \textbf{Continuous-Time Tokenization}, which directly encodes raw observation events into event-level tokens by incorporating observed values, variable identities, timestamps, inter-event intervals, and token-type indicators. Long natural gaps are explicitly modeled using GAP-aware Event tokens, while masked historical segments are replaced with MASK tokens during training, preserving the structural semantics of irregular sampling. The resulting continuous-time tokens are then processed by a standard Transformer backbone to capture long-range temporal dependencies. On the output side, we propose a simulation-free training paradigm for \textbf{Residual Flow Matching}. The model first produces a deterministic base prediction and then learns a conditional residual vector field relative to this base prediction. During training, WrapFlow directly regresses the target vector field over flow time, avoiding numerical-solver simulation and backpropagation. During inference, a small number of fixed-step Euler rollouts is sufficient to generate smooth residual correction trajectories, which are combined with the base prediction to produce the final prediction. Extensive experiments demonstrate that WrapFlow achieves state-of-the-art performance on multiple IMTS datasets. Our contributions are summarized as follows:

\begin{itemize}
    \item We propose WrapFlow, a unified continuous-time modeling framework for irregular multivariate time series forecasting, which models continuous-time dynamics from both the input and output perspectives.
    \item Technically, WrapFlow introduces Continuous-Time Tokenization to encode irregular observations without discretization and Residual Flow Matching to learn continuous residual dynamics without numerical-solver simulation, enabling efficient continuous-time forecasting.
    \item Experiments on multiple datasets show that WrapFlow outperforms state-of-the-art baselines. All datasets and code are provided in the Supplementary Material.
\end{itemize}

\section{Related Work}
\label{sec:related}

\subsection{Irregular Multivariate Time Series Forecasting}
\label{sec:rw_imtsf}

IMTSF is a critical task in domains such as clinical medicine, biomechanics, and meteorology \citep{Zhang2022Raindrop,Zhang2024TPatchGNN,Luo2025HiPatch}.
Early research primarily focused on continuous-time models, such as Neural Ordinary Differential Equations \citep{chen2018neural}, which parameterize the derivative of hidden states with neural networks and compute outputs through black-box ODE solvers, enabling continuous-time modeling.
Recently, the field has shifted toward more flexible architectures.
Set-based models like SeFT \citep{horn2020set} formulate irregularly sampled and asynchronous time series with unaligned measurements as differentiable set functions, enabling scalable modeling without requiring fixed temporal alignment.
Graph-based methods, such as Raindrop \citep{Zhang2022Raindrop}, represent each sample as a sensor graph and learn time-varying dependencies among sensors through message passing over latent graph structures.
More recently, patching mechanisms have been adapted for IMTSF.
For example, tPatchGNN \citep{Zhang2024TPatchGNN} transforms each univariate irregular time series into a series of transformable patches with uniform temporal resolution, and further models dynamic inter-series correlations with time-adaptive graph neural networks.

\subsection{Flow Matching Techniques for Time Series Analysis}
\label{sec:related_flow_matching}

Flow matching has recently become an effective framework for learning continuous transport dynamics between simple source distributions and data distributions.
By directly regressing a time-dependent vector field along predefined probability paths, it enables stable ODE-based generation without simulating a full reverse diffusion process during training \citep{lipman2022flow,tong2023improving}.
Related rectified-flow methods further improve sampling efficiency by learning straighter transport paths \citep{liu2022flow}. In time-series analysis, diffusion and score-based models have been used for probabilistic forecasting and imputation \citep{rasul2021autoregressive,tashiro2021csdi}, while FM-TS applies flow matching to time-series generation \citep{hu2024flowts}. More recent work begins to adapt flow matching to downstream tasks: TSFlow \citep{KolloviehLLSG25} combines flow matching with Gaussian-process priors for probabilistic forecasting, and CGFM \citep{xu2025bridging} models prediction residuals with conditional guided flow matching to enhance forecasting accuracy. However, these methods still mainly target sequence-level generation, forecasting, or imputation, rather than irregular multivariate forecasting with asynchronous variable-time events. Unlike previous approaches, WrapFlow uses flow matching as a query-conditioned residual refinement mechanism for irregular forecasting. Instead of generating the whole target sequence from noise, WrapFlow first predicts an anchor value and then learns a conditional residual vector field from the anchor to the ground truth.

\section{Methodology}
\label{sec:method}

\subsection{Problem Definition}
\label{sec:method_problem}

An Irregular Multivariate Time Series (IMTS) $\mathcal{O}$ is defined as a collection of $N$ univariate variables
$\{o^n_{1:L_n}\}_{n=1}^{N}$.
Each variable $o^n_{1:L_n}$ consists of $L_n$ observation tuples
$\{(t_i^n,x_i^n)\}_{i=1}^{L_n}$, where $t_i^n\in\mathbb{R}^{+}$ denotes the timestamp and $x_i^n\in\mathbb{R}$ denotes the corresponding observed value for the $i$-th observation of variable $n$.
The variables in an IMTS are sampled irregularly: sampling intervals are non-uniform within each variable, and observations are asynchronous across variables.

The IMTS forecasting task aims to predict future values from irregular multivariate historical observations $\mathcal{O}$.
Given a set of forecasting queries $\mathcal{Q}=\{[q_j^n]_{j=1}^{Q_n}\}_{n=1}^{N}$, each query $q_j^n=(s_j^n,n)$ asks for the value of variable $n$ at the future timestamp $s_j^n$.
The corresponding targets are denoted by $\mathcal{Y}=\{[y_j^n]_{j=1}^{Q_n}\}_{n=1}^{N}$.
The goal is to learn a forecasting function $\mathcal{F}_{\Theta}$ that maps historical observations and forecasting queries to future predictions:
\begin{equation}
\widehat{\mathcal{Y}}
=
\mathcal{F}_{\Theta}(\mathcal{O},\mathcal{Q})
=
\left\{
[\hat y_j^n]_{j=1}^{Q_n}
\right\}_{n=1}^{N},
\label{eq:forecast_problem}
\end{equation}
where $\Theta$ denotes the learnable parameters of WrapFlow.
For later notation, we define the valid historical observation positions and valid future target positions as
\begin{equation}
\begin{aligned}
\Omega_x
&=
\bigl\{(k,n)\mid (t_k^n,x_k^n)\text{ is a valid historical observation}\bigr\},\\
\Omega_y
&=
\bigl\{(j,n)\mid y_j^n\text{ is a valid future target}\bigr\}.
\end{aligned}
\label{eq:valid_sets}
\end{equation}

\subsection{Framework Overview}
\label{sec:method_overview}

\begin{figure*}[t]
    \centering
    \includegraphics[width=\linewidth]{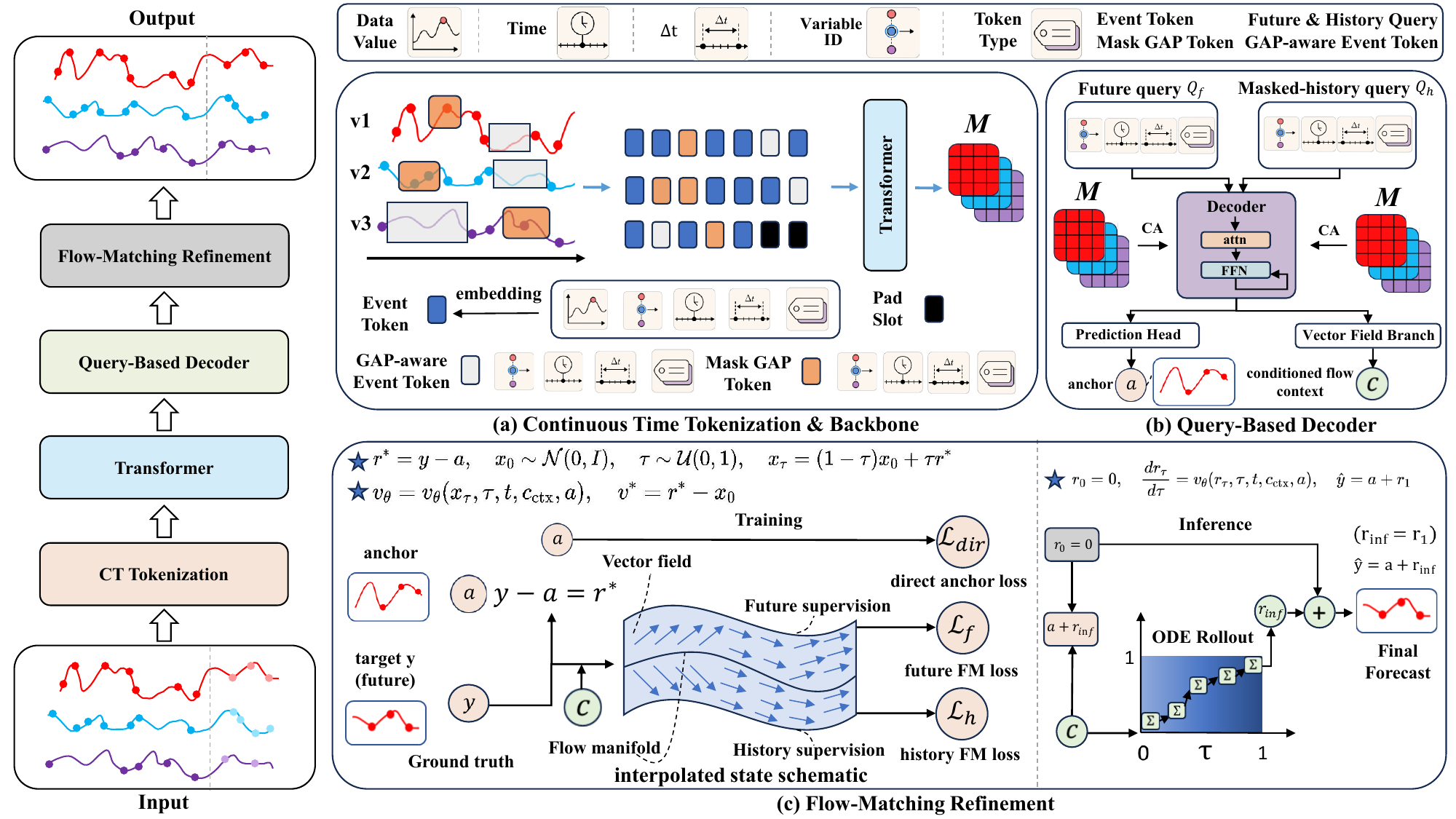}
    \caption{
    WrapFlow architecture.
    (a) Continuous-Time Tokenization and Backbone convert irregular observations into typed continuous-time tokens and encode them into variable-wise memory.
    (b) Query-Based Decoder maps value-free future and masked-history queries to temporal context for generating base future predictions and base historical reconstructions.
    (c) Flow-Matching Refinement directly regresses the target vector field over flow time during training, and during inference, it produces smooth residual correction trajectories via a small number of fixed-step Euler rollouts.
    }
    \label{fig:WrapFlow_overview}
\end{figure*}

With the forecasting task defined above, WrapFlow realizes $\mathcal{F}_{\Theta}$ through a continuous-time modeling framework.
As shown in Figure~\ref{fig:WrapFlow_overview}, WrapFlow first applies \textit{Continuous-Time Tokenization} to raw irregular observations, treating each observation event as the basic unit and encoding its value, variable identity, timestamp, and elapsed time into an event-level token.
Long unobserved intervals are explicitly marked by gap-aware tokens, and training-time masked spans are represented by Mask GAP Tokens to provide recovery supervision.
The resulting token streams are then encoded by the \textit{Transformer Backbone} into historical memory representations.
On the output side, WrapFlow constructs unified queries for future timestamps and masked historical positions, and uses a shared \textit{Query-Based Decoder} to interact with the historical memory.
Based on the decoded context, the Prediction Head first generates a deterministic anchor prediction, while the Vector Field Branch models the residual flow relative to this anchor.
During training, flow matching directly supervises the vector field; during inference, a lightweight ODE Rollout produces the residual correction, which is added to the anchor to obtain the final forecast.
We next describe these components in order.

\subsection{Continuous-Time Tokenization}
\label{sec:method_ct_tokenization}

Motivated by the irregular nature of time series discussed in Section~\ref{sec:Introduction}, we formulate IMTSF as a continuous-time modeling problem. To this end, we propose \textbf{Continuous-Time Tokenization} at the input stage to model irregular observations in continuous time. The core idea is that the model should not only capture observed values, but also reason about when events occur and the time intervals between them. Specifically, \textit{WrapFlow} encodes historical observations into three types—\textit{Event Tokens}, \textit{GAP-aware Event Tokens}, and \textit{Mask GAP Tokens}—each corresponding to a distinct semantic state in irregular histories. First, raw observations are encoded as \textit{Event Tokens}.
Second, when an observation follows a long unobserved interval, the interval itself carries meaningful semantics and is explicitly represented by a \textit{GAP-aware Event Token}. Third, during training, \textit{WrapFlow} masks a continuous historical span for recovery supervision; these positions are represented by \textit{Mask GAP Tokens}, enabling the model to distinguish natural long gaps from artificially masked segments. In summary, the three token types correspond to observed events, observations following long gaps, and masked historical positions, respectively. In the following, we introduce the three token types one by one.

\noindent\textbf{(I) Event Token.}
For variable $n$, let $(x_k^n, t_k^n)$ denote its $k$-th observed event in the historical window, and let the inter-event time gap be $\Delta t_k^n = t_k^n - t_{k-1}^n$ with $t_0^n = 0$.
WrapFlow encodes each ordinary observation event as:
\begin{equation}
\mathbf{z}_k^n
=
g_v(x_k^n)
\!+\!
\mathbf{e}_{\mathrm{var}}^{n}
\!+\!
\phi_{\mathrm{abs}}(t_k^n)
\!+\!
\phi_{\mathrm{rel}}(\Delta t_k^n)
\!+\!
\mathbf{e}_{\mathrm{type}}^{\mathrm{event}}
\!\in\!
\mathbb{R}^{d},
\label{eq:event_token}
\end{equation}
where, $g_v(\cdot)$ is the value projection, $\mathbf{e}_{\mathrm{var}}^{n}$ is the variable embedding, $\phi_{\mathrm{abs}}(\cdot)$ and $\phi_{\mathrm{rel}}(\cdot)$ are the absolute-time and relative-time encoders, and $\mathbf{e}_{\mathrm{type}}^{\mathrm{event}}$ is the learnable type embedding for ordinary observation events, $d$ denotes the hidden dimension.

\noindent\textbf{(II) GAP-aware Event Token.}
When the elapsed time since the previous visible event of the same variable exceeds a threshold $\tau_{\mathrm{gap}}$, Continuous-Time Tokenization augments the current event token with a GAP-type embedding:
\begin{equation}
\widetilde{\mathbf{z}}_k^n
=
\mathbf{z}_k^n
+
\mathbf{e}_{\mathrm{type}}^{\mathrm{gap}} \in \mathbb{R}^{d},
\label{eq:gap_token}
\end{equation}
where $\mathbf{e}_{\mathrm{type}}^{\mathrm{gap}}$ is the learnable GAP-type embedding.

\noindent\textbf{(III) Mask GAP Token.}
During training, WrapFlow samples a continuous span from the observed history for masked recovery.
For a masked event at $(x_k^n, t_k^n)$, Continuous-Time Tokenization removes its observed value and replaces the original event token with a value-free Mask GAP Token:
\begin{equation}
\mathbf{m}_k^n
=
\mathbf{e}_{\mathrm{var}}^{n}
+
\phi_{\mathrm{abs}}(t_k^n)
+
\phi_{\mathrm{rel}}(\Delta t_k^n)
+
\mathbf{e}_{\mathrm{type}}^{\mathrm{mask}}\in \mathbb{R}^{d}.
\label{eq:mask_gap_token}
\end{equation}
Here, $\mathbf{e}_{\mathrm{type}}^{\mathrm{mask}}$ is the learnable MASK-GAP-type embedding.
Compared with an Event Token, the Mask GAP Token preserves variable identity and temporal location while removing the value projection term $g_v(x_k^n)$.

\noindent\textbf{Token Sequence Construction.}
For each variable $n$, Continuous-Time Tokenization arranges its historical tokens in chronological order and pads the sequence to the maximum history length within the batch, since the number of observations varies across variables in irregular time series:
\begin{equation}
\mathbf{E}^{n}
=
\left[
\mathbf{u}_1^n,
\mathbf{u}_2^n,
\ldots,
\mathbf{u}_{K_h}^n
\right]\in\mathbb{R}^{K_h\times d},
\label{eq:padded_token_sequence}
\end{equation}
where $\mathbf{u}_k^n$ denotes the $k$-th historical token of variable $n$, and $K_h$ is the padded sequence length.
Each valid token is either an Event Token, a GAP-aware Event Token, or a Mask GAP Token.
Padding positions are masked in attention and do not carry event semantics.

\subsection{Transformer Backbone}
After Continuous-Time Tokenization, WrapFlow feeds each variable-wise token sequence into a shared backbone encoder to obtain contextual historical memory.
Padding positions are masked in attention and do not contribute to the encoded representation.
For each variable $n$, we have
\begin{equation}
\begin{aligned}
\mathbf{H}^{n}
&=
\operatorname{Backbone}(\mathbf{E}^{n})
\in\mathbb{R}^{K_h\times d},\\
\mathcal{M}
&=
\{\mathbf{H}^{n}\}_{n=1}^{N}
\in\mathbb{R}^{N \times K_h\times d}.
\end{aligned}
\label{eq:backbone_memory}
\end{equation}
where $\mathbf{H}^{n}$ denotes the contextual memory of variable $n$, $d$ is the hidden dimension, and $\mathcal{M}$ collects the memories of all variables.
WrapFlow also records the last visible historical timestamp $t_{\mathrm{last}}^n$ of each variable for subsequent use. \textit{Note that, in \textit{WrapFlow}, we adopt a standard masked attention mechanism as the backbone.}

\subsection{Query-based Decoder}
\label{sec:method_query_decoder}
Given the variable-wise historical memory $\mathcal{M}$, WrapFlow uses a shared query-based decoder to retrieve time-specific context from each variable stream.
Each query is value-free and encodes only variable identity, temporal information, and a query-type embedding.
We consider two query sets: future queries $\mathcal{Q}_f$ for forecasting and masked-history queries $\mathcal{Q}_h$ for training-time history recovery.
Both query sets share the same decoder.

\noindent\textbf{Future query $Q_f$.}
For variable $n$, let $s_j^n$ denote the $j$-th future timestamp to be predicted. Since the target value at $s_j^n$ is unknown, the query token contains no value term and is constructed as:
\begin{equation}
\mathbf{q}_{f,j}^{n}
=
\mathbf{e}_{\mathrm{var}}^{n}
+
\phi_{\mathrm{abs}}(s_j^n)
+
\phi_{\mathrm{rel}}(\Delta s_j^n)
+
\mathbf{e}_{\mathrm{type}}^{\mathrm{future}} \in \mathbb{R}^{d},
\label{eq:future_query_token}
\end{equation}
where $\Delta s_j^n$ is the elapsed time from the latest available timestamp before $s_j^n$, namely $t_{\mathrm{last}}^n$ for the first future query and $s_{j-1}^n$ for subsequent ones. The query context is then obtained by:
\begin{equation}
\mathbf{c}_{f,j}^{n}
=
\text{Cross-Attention}
\left(
\mathbf{q}_{f,j}^{n},
\mathbf{H}^{n}, \mathbf{H}^{n}
\right).
\label{eq:future_decoder}
\end{equation}
The resulting context $\mathbf{c}_{f,j}^{n}$ is used for base prediction and subsequent residual flow refinement.

\noindent\textbf{Masked-history query $Q_h$.}
During training, WrapFlow samples a continuous span of observed historical events, yielding a masked set $\Omega_h\subseteq\Omega_x$.
For each masked event $(x_k^n,t_k^n)\in\Omega_h$, we construct a value-free history query at the same variable and timestamp.
Accordingly, the masked-history query set is
$\mathcal{Q}_h=\{(t_k^n,n)\mid (x_k^n,t_k^n)\in\Omega_h\}$,
and each query token is defined as
\begin{equation}
\begin{aligned}
\mathbf{q}_{h,k}^{n}
&=
\mathbf{e}_{\mathrm{var}}^{n}
+
\phi_{\mathrm{abs}}(t_k^n)
+
\phi_{\mathrm{rel}}(\Delta t_k^n)
+
\mathbf{e}_{\mathrm{type}}^{\mathrm{history}},\\
\mathbf{c}_{h,k}^{n}
&=
\text{Cross-Attention}
\left(
\mathbf{q}_{h,k}^{n},
\mathbf{H}^{n}, \mathbf{H}^{n}
\right).
\end{aligned}
\label{eq:history_query_token}
\end{equation}
where $\mathbf{e}_{\mathrm{type}}^{\mathrm{history}}$ is the learnable history-query type embedding.
The masked value $x_k^n$ serves as the supervision target for training-time history recovery.

\noindent\textbf{Base prediction and reconstruction.}
Given the decoder context, WrapFlow first produces a deterministic base estimate through a shared MLP prediction head.
For future queries, this estimate serves as the initial prediction of the target future value; for masked-history queries, it serves as the reconstruction of the masked historical observation.
Concretely, the two outputs are computed as
\begin{equation}
\begin{aligned}
\bar{x}_{f,j}^n
&=
\mathrm{MLP}
\left(
[\mathbf{c}_{f,j}^{n}\parallel \psi(s_j^n)]
\right),\\
\bar{x}_{h,k}^n
&=
\mathrm{MLP}
\left(
[\mathbf{c}_{h,k}^{n}\parallel \psi(t_k^n)]
\right).
\end{aligned}
\label{eq:base_predictions}
\end{equation}
where $\psi(\cdot)$ denotes the time embedding used in the prediction head.
The residual flow then models only the correction from the base estimate to the target value.

\subsection{Flow-Matching Refinement}
\label{sec:method_flow_matching}

After obtaining the base estimate, WrapFlow does not directly use it as the final prediction, but further refines it in residual space.

To strengthen residual refinement by supervising the vector field with both future targets and irregular historical context, WrapFlow applies flow matching to two complementary branches: a \textbf{Future branch} for the main forecasting task and a \textbf{Masked-history branch} for training-time auxiliary self-supervision.

Specifically, in the Future branch the final prediction $y_j^n$ is decomposed into the base estimate and a residual:
$y_j^n = \bar{x}_{f,j}^n + r_j^{n},$
where $r_{j}^{n}$ is the difference between the true value and the base estimate, and serves as the target for flow matching. For masked-history recovery, we analogously define the residual target as $r_{h,k}^{n}=x_k^n-\bar{x}_{h,k}^{n}$.
When constructing this target, $r_{j}^{n}$ is detached from the base-estimate branch, so the flow loss mainly optimizes the residual vector field.

\label{sec:method_residual_fm}
\par\medskip
\noindent\textbf{Future branch.}\enspace
Flow matching in residual field learns how to correct the base-estimate error.
To make the prediction more accurate, instead of directly regressing the final residual, it trains the model to predict the distribution from random noise $x_{0,j}^{n}\sim\mathcal{N}(0,I)$ toward the target residual.
In the training stage, for each valid future target $(j,n)\in\Omega_y$, the model samples a flow step $\tau\sim\mathcal{U}(0,1)$.
Here, $\tau$ is only an artificial path parameter for flow matching.

The model linearly interpolates between the initial noise and the target residual:
\begin{equation}
x_{\tau,j}^{n}
=
(1-\tau)x_{0,j}^{n}
+
\tau r_{j}^{n},
\qquad
v_{\tau,j}^{n}
=
r_{j}^{n}
-
x_{0,j}^{n}.
\label{eq:interpolation_velocity}
\end{equation}
Here, $x_{\tau,j}^{n}$ is a residual-space state, and $v_{\tau,j}^{n}$ is the target velocity along the interpolation path.

The Vector Field Branch predicts the correction direction conditioned on the current residual state $x_{\tau,j}^{n}$, flow step $\tau$, physical target time $s_j^n$, query context $\mathbf{c}_{f,j}^{n}$, and base estimate $\bar{x}_{f,j}^n$ :
\begin{equation}
\hat{v}_{\tau,j}^{n}
=
v_{\theta}
\left(
x_{\tau,j}^{n},
\tau,
s_j^n,
\mathbf{c}_{f,j}^{n},
\bar{x}_{f,j}^n
\right),
\label{eq:vector_field}
\end{equation}
where $\hat{v}_{\tau,j}^{n}$ denotes the predicted residual velocity for the $j$-th future query of variable $n$ at flow time $\tau$, and $v_{\theta}(\cdot)$ is the learnable scalar decoder with parameters $\theta$.

We only compute this training objective on valid future positions to supervise future residual refinement in future branch.

\par\medskip
\noindent\textbf{Masked-history branch.}\enspace
This branch uses a similar objective: for each masked position $(k,n)\in\Omega_h$, the model defines the residual target as $r_{h,k}^{n}=x_k^n-\bar{x}_{h,k}^{n}$, then constructs the corresponding interpolation state and target velocity.
After sampling $x_{0,h,k}^{n}\sim\mathcal{N}(0,I)$ and $\tau'\sim\mathcal{U}(0,1)$, the history interpolation state and target velocity are:
\begin{equation}
x_{\tau',h,k}^{n}
=
(1-\tau')x_{0,h,k}^{n}
+
\tau' r_{h,k}^{n},
\quad
v_{\tau,h,k}^{n}
=
r_{h,k}^{n}
-
x_{0,h,k}^{n}.
\label{eq:history_interpolation_velocity}
\end{equation}
The velocity $\hat{v}_{\tau,h,k}^{n}$ is also derived from the shared network $v_\theta$ of the future branch, using the masked-history context $\mathbf{c}_{h,k}^{n}$, timestamp $t_k^n$, flow step $\tau'$ and base reconstruction $\bar{x}_{h,k}^{n}$.

The losses of the two branches are as follows:
\begin{equation}
\begin{aligned}
\mathcal{L}_{f}
&=
\frac{1}{|\Omega_y|}
\sum_{(j,n)\in\Omega_y}
\ell_{\mathrm{flow}}
\left(
\hat{v}_{\tau,j}^{n},
v_{\tau,j}^{n}
\right),\\
\mathcal{L}_{h}
&=
\frac{1}{|\Omega_h|}
\sum_{(k,n)\in\Omega_h}
\ell_{\mathrm{flow}}
\left(
\hat{v}_{\tau,h,k}^{n},
v_{\tau,h,k}^{n}
\right).
\end{aligned}
\label{eq:flow_losses}
\end{equation}
where $\ell_{\mathrm{flow}}(\cdot,\cdot)$ denotes the velocity matching loss.
The history FM loss provides additional self-supervision and helps the residual vector field use irregular historical context.

\subsection{Training Objective and Inference Stage}
\label{sec:method_inference}
\par\medskip
\noindent\textbf{Training objective.}\enspace
During training, WrapFlow jointly optimizes the base estimate and residual flow refinement.
The direct prediction loss is computed only on valid future targets and trains the Prediction Head.
The future FM loss trains residual correction for future forecasting, while the history FM loss trains the Masked-history branch and regularizes the residual vector field. The history branch is used only during training as auxiliary self-supervision and is not required at inference time.
The direct prediction loss is defined as
\begin{equation}
\mathcal{L}_{\mathrm{dir}}
=
\frac{1}{|\Omega_y|}
\sum_{(j,n)\in\Omega_y}
\ell_{\mathrm{dir}}
\left(
\bar{x}_{f,j}^n,
y_j^n
\right),
\label{eq:direct_prediction_loss}
\end{equation}
where $\ell_{\mathrm{dir}}(\cdot,\cdot)$ denotes the direct forecasting loss for the base estimate.
The overall objective consists of these three parts:
\begin{equation}
\mathcal{L}
=
\mathcal{L}_{\mathrm{dir}}
+
\lambda_f \mathcal{L}_{f}
+
\lambda_h \mathcal{L}_{h}.
\label{eq:total_loss}
\end{equation}
Here, $\lambda_f$ and $\lambda_h$ control the weights of future FM loss and history FM loss, respectively.

\par\medskip
\noindent\textbf{Inference stage.}\enspace
At validation and test time, WrapFlow uses the learned vector field to generate an actual residual correction.
For each Future query, the model starts from a query-specific zero residual state $r_{0,j}^{n}=0$ and performs ODE Rollout over $\tau\in[0,1]$:
\begin{equation}
\frac{d r_{\tau,j}^{n}}{d\tau}
=
v_{\theta}
\left(
r_{\tau,j}^{n},
\tau,
s_j^n,
\mathbf{c}_{f,j}^{n},
\bar{x}_{f,j}^n
\right),
\;\tau\in[0,1],
\;r_{0,j}^{n}=0.
\label{eq:inference_ode}
\end{equation}
The terminal state $r_{1,j}^{n}$ is the final residual correction, which is added to the base estimate to obtain the prediction:
\begin{equation}
\hat{y}_j^n
=
\bar{x}_{f,j}^n
+
r_{1,j}^{n}.
\label{eq:final_forecast}
\end{equation}
In practice, the ODE Rollout can be performed with lightweight Euler integration.

\section{Experiments}
\label{sec:experiments}

In Section~\ref{sec:exp_settings}, we introduce the datasets, baselines, and implementation details. Section~\ref{sec:main_results} presents the main experimental results. Section~\ref{sec:Ablation_Study_and_Analysis} presents ablation studies of key components, and Section~\ref{sec:parameter_sensitivity} studies the effect of the hidden dimension of WrapFlow.

Due to space constraints, extended lookback and extended forecast-horizon experiments are deferred to Sections A.4 and A.5 of the Supplementary Material, respectively. Additional sensitivity results for encoder depth are provided in Section A.6.

\begin{table*}[t]
    \centering
    {\small
    \setlength{\tabcolsep}{.4pt}
    \renewcommand{\arraystretch}{1.26}
    \begin{tabular*}{\textwidth}{@{\extracolsep{\fill}}l*{8}{c}@{}}
        \toprule
        \textbf{Dataset} & \multicolumn{2}{c}{\textbf{HumanActivity}} & \multicolumn{2}{c}{\textbf{USHCN}} & \multicolumn{2}{c}{\textbf{PhysioNet}} & \multicolumn{2}{c}{\textbf{MIMIC}} \\
        \midrule
        \textbf{Metrics}  & \textbf{MSE}           & \textbf{MAE}           & \textbf{MSE}         & \textbf{MAE}         & \textbf{MSE}         & \textbf{MAE}         & \textbf{MSE}         & \textbf{MAE}         \\
        \midrule
        PrimeNet    & 4.2507$\pm$.0041 & 1.7018$\pm$.0011 & .4930$\pm$.0015 & .4954$\pm$.0018 & .7953$\pm$.0000 & .6859$\pm$.0001 & .9073$\pm$.0001 & .6614$\pm$.0001 \\
        NeuralFlows & .1722$\pm$.0090 & .3150$\pm$.0094 & .2087$\pm$.0258 & .3157$\pm$.0187 & .4056$\pm$.0033 & .4466$\pm$.0027 & .6085$\pm$.0101 & .5306$\pm$.0066 \\
        CRU         & .1387$\pm$.0073 & .2607$\pm$.0092 & .2168$\pm$.0162 & .3180$\pm$.0248 & .6179$\pm$.0045 & .5778$\pm$.0031 & .5895$\pm$.0092 & .5151$\pm$.0048 \\
        mTAN        & .0993$\pm$.0026 & .2219$\pm$.0047 & .5561$\pm$.2020 & .5015$\pm$.0968 & .3809$\pm$.0043 & .4291$\pm$.0035 & .9408$\pm$.1126 & .6755$\pm$.0459 \\
        SeFT        & 1.3786$\pm$.0024 & .9762$\pm$.0007 & .3345$\pm$.0022 & .4083$\pm$.0084 & .7721$\pm$.0021 & .6760$\pm$.0029 & .9230$\pm$.0015 & .6628$\pm$.0008 \\
        GNeuralFlow & .3936$\pm$.1585 & .4541$\pm$.0841 & .2205$\pm$.0421 & .3286$\pm$.0412 & .8207$\pm$.0310 & .6759$\pm$.0100 & .8957$\pm$.0209 & .6450$\pm$.0072 \\
        GRU-D       & .1893$\pm$.0627 & .3253$\pm$.0485 & .2097$\pm$.0493 & .3045$\pm$.0305 & .3419$\pm$.0029 & .3992$\pm$.0011 & .4759$\pm$.0100 & .4526$\pm$.0055 \\
        Raindrop    & .0916$\pm$.0072 & .2114$\pm$.0072 & .2035$\pm$.0336 & .3029$\pm$.0264 & .3478$\pm$.0019 & .4044$\pm$.0020 & .6754$\pm$.1829 & .5444$\pm$.0868 \\
        Warpformer  & .0449$\pm$.0010 & .1228$\pm$.0018 & .1888$\pm$.0598 & .2939$\pm$.0591 & \underline{.3056$\pm$.0011} & .3661$\pm$.0016 & .4302$\pm$.0035 & .4025$\pm$.0014 \\
        tPatchGNN   & .0443$\pm$.0009 & .1247$\pm$.0031 & .1885$\pm$.0403 & .3084$\pm$.0479 & .3133$\pm$.0053 & .3697$\pm$.0049 & .4431$\pm$.0115 & .4077$\pm$.0088 \\
        GraFITi     & .0437$\pm$.0005 & .1221$\pm$.0017 & .1691$\pm$.0093 & .2777$\pm$.0248 & .3075$\pm$.0015 & \underline{.3637$\pm$.0036} & .4359$\pm$.0455 & .4142$\pm$.0297 \\
        Hi-Patch    & .0435$\pm$.0002 & .1204$\pm$.0009 & .1749$\pm$.0268 & .2717$\pm$.0216 & .3071$\pm$.0029 & .3675$\pm$.0042 & \underline{.4279$\pm$.0010} & .4033$\pm$.0032 \\
        KAFNet      & \underline{.0429$\pm$.0003} & .1161$\pm$.0010 & .1698$\pm$.0181 & .2690$\pm$.0226 & .3164$\pm$.0028 & .3715$\pm$.0038 & .4402$\pm$.0086 & .4102$\pm$.0041 \\
        APN         & \textbf{.0421$\pm$.0001} & \underline{.1159$\pm$.0006} & \underline{.1590$\pm$.0137} & \underline{.2611$\pm$.0167} & .3093$\pm$.0011 & .3650$\pm$.0026 & .4292$\pm$.0027 & \underline{.4016$\pm$.0016} \\
        \midrule
        \textbf{WrapFlow} & .0449$\pm$.0007 & \textbf{.1154$\pm$.0009} & \textbf{.1401$\pm$.0191} & \textbf{.2233$\pm$.0054} & \textbf{.3047$\pm$.0003} & \textbf{.3555$\pm$.0055} & \textbf{.4218$\pm$.0031} & \textbf{.3838$\pm$.0038} \\
        \bottomrule
    \end{tabular*}
    }
    \caption{Forecasting performance on four IMTS datasets. Overall performance is evaluated by MSE and MAE (mean $\pm$ std). The best and second-best results are highlighted in \textbf{bold} and with an \underline{underline}, respectively. Leading zeros are omitted for all decimal values.}
    \label{tab:overall_performance_adjusted}
\end{table*}

\subsection{Experimental Settings}
\label{sec:exp_settings}

\subsubsection{Datasets}

\begin{table}[t]
\centering
{\small
\setlength{\tabcolsep}{2.2pt}
\renewcommand{\arraystretch}{1.08}
\begin{tabular*}{\columnwidth}{@{\extracolsep{\fill}}lcccc@{}}
\toprule
\textbf{Dataset} & \textbf{\# Vars.} & \textbf{\# Samples} & \textbf{Avg. \# Obs.} & \textbf{Max Len.} \\
\midrule
PhysioNet        & 36 & 11,981 & 308.6 & 47 \\
MIMIC            & 96 & 21,250 & 144.6 & 96 \\
HumanActivity    & 12 & 1,359 & 362.2 & 131 \\
USHCN            & 5 & 1,114 & 313.5 & 337 \\
\bottomrule
\end{tabular*}
}
\caption{Statistics of the four IMTS datasets used in our experiments.}
\label{tab:dataset_statistics}
\end{table}

To ensure comprehensive and fair comparisons across different methods, we evaluate WrapFlow on four widely used benchmarks for irregular multivariate time series forecasting: PhysioNet, MIMIC, HumanActivity, and USHCN.
These datasets cover diverse real-world domains, including healthcare,biomechanics, and climate science. Their detailed statistics are summarized in Table~\ref{tab:dataset_statistics}.

\subsubsection{Baselines}

We comprehensively evaluate WrapFlow against 14 baselines:
1) IMTS classification/imputation models, comprising PrimeNet \citep{Chowdhury2023PrimeNet}, SeFT \citep{horn2020set}, mTAN \citep{MTAN}, GRU-D \citep{che2018recurrent}, Raindrop \citep{Zhang2022Raindrop}, and Warpformer \citep{Zhang2023Warpformer};
and 2) IMTS forecasting models, which include NeuralFlows \citep{bilovs2021neural}, CRU \citep{schirmer2022modeling}, GNeuralFlow \citep{mercatali2024graph}, tPatchGNN \citep{Zhang2024TPatchGNN}, GraFITi \citep{Yalavarthi2024GraFITi}, Hi-Patch \citep{Luo2025HiPatch}, KAFNet \citep{zhou2026revitalizing}, and APN \citep{liu2026rethinking}.
These baselines cover representative continuous-time, recurrent, set-based, graph-based, patch-based, and recent state-of-the-art IMTS forecasting architectures.

\subsubsection{Implementation Details}

The lookback time periods are 36 hours for MIMIC-III and PhysioNet, 3000 milliseconds for HumanActivity, and 3 years for USHCN.
HumanActivity uses 300 milliseconds as the forecast length, while the remaining datasets use the next 3 timestamps as forecast targets, following the settings in existing works \citep{liu2026rethinking,li2025hyperimts,bilovs2021neural,de2019gru}.
Following previous IMTSF studies, we adopt Mean Squared Error (MSE) and Mean Absolute Error (MAE) as evaluation metrics.
To ensure reproducibility and mitigate the effects of randomness, each experiment is run independently with five different random seeds from 2024 to 2028, and we report the mean and standard deviation. Following the fairness protocols of TFB and TAB~\citep{qiu2024tfb,qiu2025tab}, we do not use the ``Drop Last'' trick during evaluation. All experiments are implemented in PyTorch 2.6.0+cu124~\citep{paszke2019pytorch} and run on an NVIDIA Tesla A800 GPU.

\subsection{Main Results}
\label{sec:main_results}

Comprehensive results are presented in Table~\ref{tab:overall_performance_adjusted} to demonstrate the performance of WrapFlow.
We have the following observations:
1) Compared with a wide range of forecasting models, WrapFlow achieves superior predictive performance, attaining the best results on most evaluation metrics.
Notably, it outperforms the second-best method, APN, by 3\% in MSE and 5.7\% in MAE. These gains indicate that modeling continuous-time dynamics provides substantial benefits.
2) WrapFlow delivers consistently strong performance across datasets from diverse application domains—ranging from clinical time series (PhysioNet and MIMIC) to biomechanics (HumanActivity) and climate science (USHCN). This consistency indicates that the proposed continuous-time modeling framework captures temporal patterns across datasets with different characteristics.

\begin{figure*}[t]
    \centering
    \includegraphics[width=0.89\textwidth]{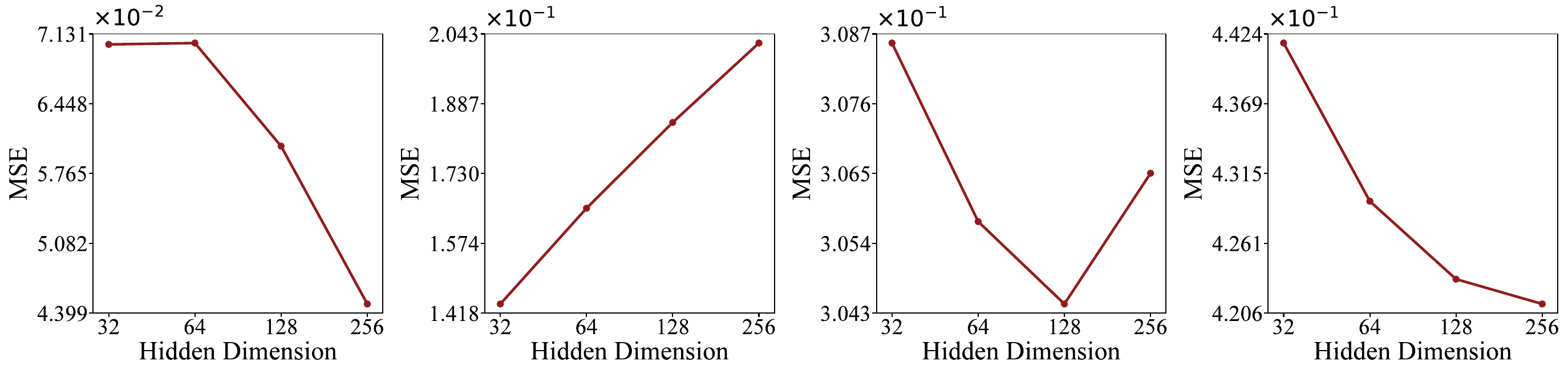}\\
    \caption{Parameter sensitivity of WrapFlow with respect to the hidden dimension $D$ on HumanActivity, USHCN, PhysioNet, and MIMIC. Each subfigure reports the MSE under different hidden dimensions.}
    \label{fig:main_param_sensitivity}
\end{figure*}

\begin{table*}[t]
    \centering
    {\small
    \setlength{\tabcolsep}{0.45pt}
    \renewcommand{\arraystretch}{1.24}
    \begin{tabular*}{\textwidth}{
        @{\extracolsep{\fill}}
        >{\centering\arraybackslash}m{0.17\textwidth}
        cccccccc
        @{}
    }
        \hline
        \textbf{Dataset}
        & \multicolumn{2}{c}{\textbf{HumanActivity}}
        & \multicolumn{2}{c}{\textbf{USHCN}}
        & \multicolumn{2}{c}{\textbf{PhysioNet}}
        & \multicolumn{2}{c}{\textbf{MIMIC}} \\
        \hline
        \textbf{Metrics}
        & \textbf{MSE} & \textbf{MAE}
        & \textbf{MSE} & \textbf{MAE}
        & \textbf{MSE} & \textbf{MAE}
        & \textbf{MSE} & \textbf{MAE} \\
        \hline
        \makecell{w/o Continuous-Time\\Tokenization}
        & .0704$\pm$.0003 & .1585$\pm$.0003
        & .1555$\pm$.0093 & .2377$\pm$.0090
        & .3968$\pm$.0008 & .4201$\pm$.0007
        & .4805$\pm$.0024 & .4167$\pm$.0021 \\

        \midrule
        \makecell{w/o Flow-Matching\\Refinement}
        & .0506$\pm$.0075 & .1241$\pm$.0129
        & .1668$\pm$.0300 & .2597$\pm$.0095
        & .3066$\pm$.0016 & .3600$\pm$.0038
        & .4278$\pm$.0021 & .3961$\pm$.0030 \\

        \midrule
        w/o Cross-Attention
        & .0455$\pm$.0011 & .1178$\pm$.0016
        & .1501$\pm$.0114 & .2279$\pm$.0108
        & .3053$\pm$.0013 & .3555$\pm$.0036
        & .4260$\pm$.0027 & .3855$\pm$.0013 \\

        \midrule
        \makecell{w/o Gap-aware\\Event Token}
        & .0461$\pm$.0017 & .1157$\pm$.0015
        & .1519$\pm$.0370 & .2253$\pm$.0097
        & .3056$\pm$.0006 & .3559$\pm$.0052
        & .4239$\pm$.0033 & .3845$\pm$.0039 \\
    
        \midrule
        \makecell{w/o Relative\\Positional Encoding}
        & .0458$\pm$.0005 & .1163$\pm$.0012
        & .1437$\pm$.0120 & .2282$\pm$.0127
        & .3083$\pm$.0010 & .3581$\pm$.0054
        & .4262$\pm$.0010 & .3847$\pm$.0019 \\
        
        \midrule
        \textbf{WrapFlow (Ours)}
        & \textbf{.0449$\pm$.0007} & \textbf{.1154$\pm$.0009}
        & \textbf{.1401$\pm$.0191} & \textbf{.2233$\pm$.0054}
        & \textbf{.3047$\pm$.0003} & \textbf{.3555$\pm$.0055}
        & \textbf{.4218$\pm$.0031} & \textbf{.3838$\pm$.0038} \\
        \hline
    \end{tabular*}
    }
    \caption{Ablation studies for WrapFlow. Overall performance is evaluated
    by MSE and MAE (mean $\pm$ std). The best results are highlighted in
    \textbf{bold}. Leading zeros are omitted for all decimal values.}
    \label{tab:ablation}
\end{table*}

\subsection{Ablation Study and Analysis}
\label{sec:Ablation_Study_and_Analysis}

\textbf{Ablation study of key components in WrapFlow.}
We perform ablation studies to validate the contribution of
key components in WrapFlow—see Table~\ref{tab:ablation}. We make the following observations: 1) Removing CT Tokenization leads to the most significant performance degradation, demonstrating that typed continuous-time tokens are crucial for preserving observation values, temporal information, and token semantics in irregular time series.
2) Removing Flow-Matching Refinement and relying only on the deterministic prediction head degrades forecasting accuracy, indicating that residual flow refinement provides complementary correction beyond the anchor forecast.
3) Replacing Cross-Attention with a pooled-memory concatenation mechanism reduces performance, confirming that query-specific retrieval from historical memory is important for adaptive future conditioning.
4) Removing GAP-aware Event Token also weakens performance, suggesting that long unobserved intervals contain useful temporal information rather than merely representing missing regions.
5) Removing Relative Positional Encoding impairs accuracy, showing that elapsed-time information between irregular events is essential for modeling non-uniform temporal dynamics.

\subsection{Parameter Sensitivity}
\label{sec:parameter_sensitivity}

We study the hidden dimension $D$, which directly affects the representation capacity of WrapFlow. We evaluate its impact on forecasting performance across different datasets.

\noindent\textbf{Hidden dimension.}
Figure~\ref{fig:main_param_sensitivity} shows that the effect of the hidden dimension $D$ reflects a trade-off between representation capacity and over-parameterization. A larger $D$ allows WrapFlow to capture richer temporal patterns, which benefits more complex datasets such as HumanActivity and MIMIC. In contrast, USHCN achieves the best performance with a smaller hidden dimension, suggesting that its temporal dynamics are relatively regular and do not require a large latent space. PhysioNet achieves the best performance at a moderate scale. Overall, WrapFlow does not simply benefit from increasing model size; instead, the hidden dimension $D$ should be selected according to the intrinsic complexity of each dataset.

\section{Conclusion}
\label{sec:conclusion}

In this paper, we propose WrapFlow, a continuous-time enhancement framework for irregular multivariate time series forecasting. By introducing Continuous-Time Tokenization at the input stage and a simulation-free Residual Flow Matching mechanism at the output stage, WrapFlow effectively models continuous-time dynamics without relying on discretization or computationally
expensive numerical solvers. This design enables accurate and efficient forecasting while preserving the intrinsic temporal semantics of irregular observations. Experiments on multiple datasets show that WrapFlow outperforms SOTA baselines. All datasets and code are provided in the Supplementary Material.
\bibliography{reference,Sections/local-refs}

\clearpage
\appendix
\section{Experiments setup details}
\label{sec:appendix_datasets}

\subsection{Datasets}
\label{Appendix-Datasets}
\label{app:datasets}

We evaluate WrapFlow on four widely used irregular multivariate time series forecasting benchmarks: PhysioNet, MIMIC-III, HumanActivity, and USHCN. These datasets cover clinical monitoring, human motion, and climate observations, providing complementary sparsity patterns and temporal scales. For PhysioNet, MIMIC-III, and USHCN, we use the preprocessing protocols adopted in prior irregular forecasting studies~\citep{Yalavarthi2024GraFITi,li2025hyperimts}. For HumanActivity, we follow the setup used by Warpformer and later benchmarks~\citep{Zhang2023Warpformer,li2025hyperimts}. All datasets are split into training, validation, and test sets with an 80\%, 10\%, and 10\% ratio.

\textbf{PhysioNet.}
PhysioNet 2012~\citep{silva2012predicting} contains irregular clinical measurements collected during the first 48 hours of ICU stays. The original challenge data include 12,000 patient records with 41 clinical signals. Following the common forecasting protocol, the processed benchmark used in our experiments contains 36 variables and 11,981 samples, and each sequence is divided into observation and prediction windows.

\textbf{MIMIC-III.}
MIMIC-III~\citep{johnson2016mimic} is a large critical-care database. We use the irregular forecasting version with 96 clinical variables measured during the first 48 hours after ICU admission. After preprocessing, the benchmark contains 21,250 samples.

\textbf{HumanActivity.}
HumanActivity contains 3D positional sensor measurements from five subjects. It has 12 variables observed at irregular timestamps, and the time series are segmented into 4,000 millisecond windows. The benchmark protocol used here reports 1,359 samples after preprocessing.

\textbf{USHCN.}
USHCN~\citep{easterling2002united} provides long-term climate records from U.S. weather stations over five variables. Following standard preprocessing, we use 1,114 stations from 1996 to 2000, resulting in irregular sequences with a maximum length of 337 observations.

\subsection{Baselines}
\label{Appendix-Baselines}

\textbf{PrimeNet}~\citep{Chowdhury2023PrimeNet} pretrains representations for irregular multivariate time series and adapts them to downstream tasks.

\textbf{NeuralFlows}~\citep{bilovs2021neural} models trajectory densities with continuous-time normalizing flows, providing an efficient alternative to Neural ODE-based sequence modeling.

\textbf{CRU}~\citep{schirmer2022modeling} uses a continuous recurrent unit to evolve hidden states in continuous time and combine recurrent modeling with stochastic dynamics.

\textbf{mTAN}~\citep{MTAN} learns multi-time attention over irregular observations by mapping observations to reference time points.

\textbf{SeFT}~\citep{horn2020set} treats irregular observations as sets and applies permutation-invariant attention to reduce sensitivity to sampling frequency.

\textbf{GNeuralFlow}~\citep{mercatali2024graph} extends NeuralFlows with graph neural networks to capture interactions among irregularly sampled variables.

\textbf{GRU-D}~\citep{che2018recurrent} augments recurrent networks with trainable decay terms that encode elapsed time and missingness.

\textbf{Raindrop}~\citep{Zhang2022Raindrop} constructs graph-guided representations for irregular clinical time series and propagates information among sensors.

\textbf{Warpformer}~\citep{Zhang2023Warpformer} introduces time-warping into Transformer blocks to model non-stationary and irregular clinical sequences.

\textbf{tPatchGNN}~\citep{Zhang2024TPatchGNN} divides irregular time series into temporal patches and applies graph neural networks to learn cross-variable dependencies.

\textbf{GraFITi}~\citep{Yalavarthi2024GraFITi} formulates irregular time series forecasting with bipartite graphs between observations and target query points.

\textbf{Hi-Patch}~\citep{Luo2025HiPatch} hierarchically patches irregular sequences to jointly capture local details and global temporal patterns.

\textbf{KAFNet}~\citep{zhou2026revitalizing} combines canonical pre-alignment with frequency-domain attention for efficient irregular forecasting.

\textbf{APN}~\citep{liu2026rethinking} uses time-aware patch aggregation to regularize irregular sequences and provide a strong simple baseline.

\begin{table}[t]
\centering
{\small
\setlength{\tabcolsep}{2.2pt}
\renewcommand{\arraystretch}{1.08}
\begin{tabular*}{\columnwidth}{@{\extracolsep{\fill}}lcccc@{}}
\toprule
\textbf{Dataset} & \textbf{\# Vars.} & \textbf{\# Samples} & \textbf{Avg. \# Obs.} & \textbf{Max Len.} \\
\midrule
PhysioNet        & 36 & 11,981 & 308.6 & 47 \\
MIMIC            & 96 & 21,250 & 144.6 & 96 \\
HumanActivity    & 12 & 1,359 & 362.2 & 131 \\
USHCN            & 5 & 1,114 & 313.5 & 337 \\
\bottomrule
\end{tabular*}
}
\caption{Dataset statistics.}
\label{tab:appendix_dataset_statistics}
\end{table}

\subsection{Implementation Details}
\label{Appendix-Implementation-Details}
\label{app:implementation}

Following previous work, we report Mean Squared Error (MSE) and Mean Absolute Error (MAE) for all forecasting tasks. The lookback windows are set to 36 hours for PhysioNet and MIMIC-III, 3000 milliseconds for HumanActivity, and 3 years for USHCN. HumanActivity uses a 300 millisecond forecasting window, while the other three datasets use the next 3 timestamps as prediction targets, matching established irregular forecasting settings~\citep{liu2026rethinking,li2025hyperimts,bilovs2021neural,de2019gru}. In line with the fairness protocols of TFB and TAB~\citep{qiu2024tfb,qiu2025tab}, we do not use the ``Drop Last'' trick during evaluation. All experiments are implemented in PyTorch 2.6.0+cu124~\citep{paszke2019pytorch} and run on a server equipped with an NVIDIA Tesla A800 GPU.

\subsection{Varying Lookback Lengths}
\label{app:varying_lookback_horizons}

\begin{figure*}[t]
\centering
\includegraphics[width=\textwidth]{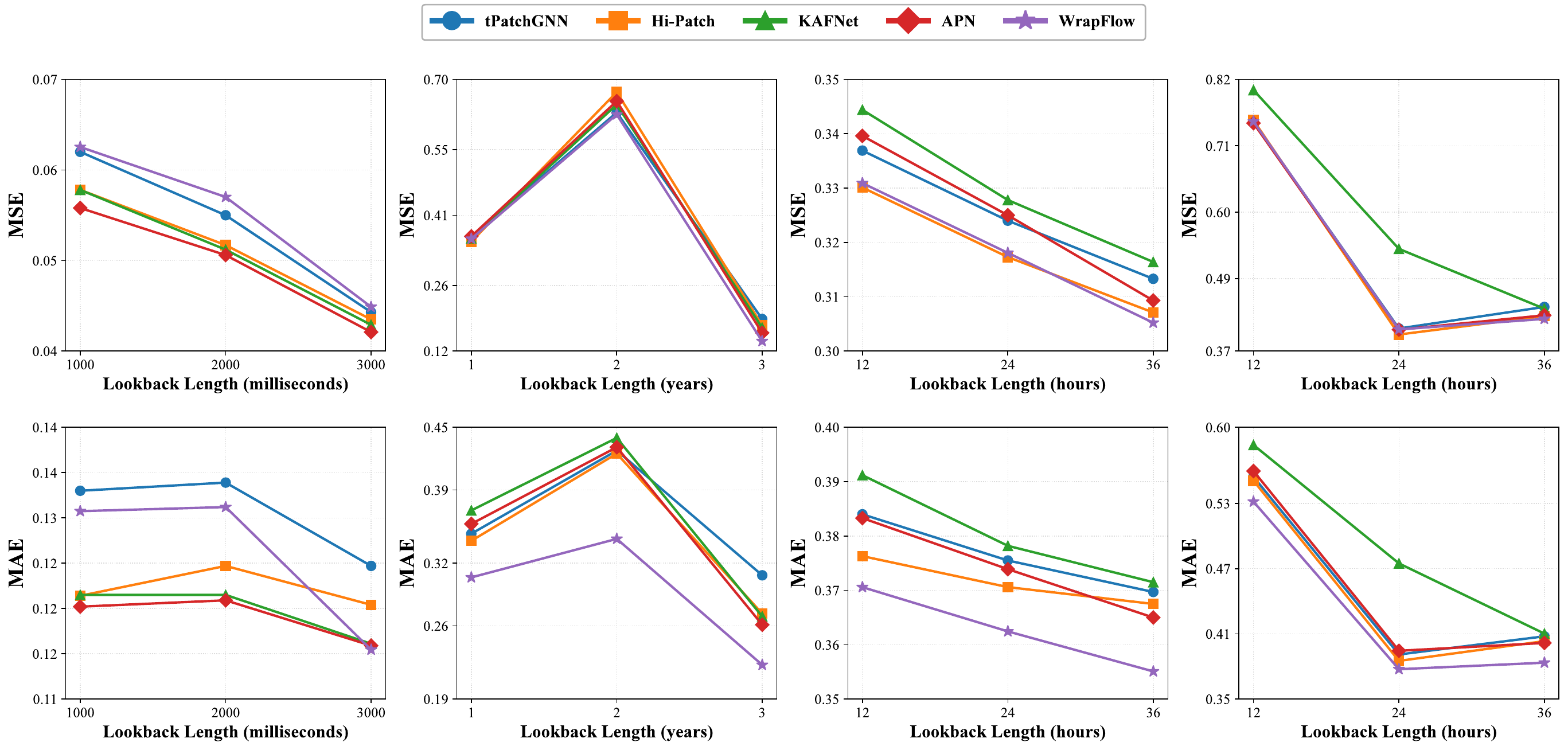}
\caption{Forecasting performance with varying lookback lengths and fixed forecast horizons. From left to right, the datasets are HumanActivity, USHCN, PhysioNet, and MIMIC; the top row reports MSE and the bottom row reports MAE.}
\label{fig:varying_lookback_length}
\end{figure*}

We further study how WrapFlow responds to different lookback lengths while keeping the forecast horizons fixed according to the main-paper experimental setting. For HumanActivity, USHCN, and PhysioNet, the lookback lengths are set to 1000/2000/3000 milliseconds, 50/100/150 observations, and 12/24/36 hours, respectively; for MIMIC, they are set to 24/48/72 hours. As shown in Figure~\ref{fig:varying_lookback_length}, WrapFlow remains stable across the tested history windows and generally benefits from richer context when the extra history is informative. The top row reports MSE and the bottom row reports MAE. The effect is most consistent on HumanActivity and PhysioNet, while USHCN and MIMIC show milder, dataset-specific fluctuations, indicating that WrapFlow can adapt to both steadily improving and saturation-like lookback patterns without large performance degradation.

\subsection{Varying Forecast Horizons}
\label{app:varying_forecast_horizons}

We further evaluate the robustness of WrapFlow under longer-term prediction by varying the forecast horizons while keeping the lookback lengths consistent with the main-paper experimental setting and comparing against competitive IMTS forecasting baselines.

Following the whole-series horizon settings adopted in prior irregular forecasting studies~\citep{liu2026rethinking,li2025hyperimts}, the forecast horizons are set to 12 hours for MIMIC-III and PhysioNet'12, 1000 milliseconds for HumanActivity, and 1 year for USHCN. The results are summarized in Table~\ref{tab:varying_forecast_horizons}. WrapFlow achieves the best performance on USHCN, PhysioNet, and MIMIC across both MSE and MAE, while remaining competitive on HumanActivity under the longest horizon setting. These results indicate that WrapFlow remains robust when the prediction horizon is extended. We attribute this stability to its continuous-time conditioning and flow-based refinement, which help preserve globally consistent temporal dynamics even when local short-term cues become less reliable.

\begin{table*}[t]
\centering
{\small
\setlength{\tabcolsep}{1.7pt}
\begin{tabular*}{\textwidth}{@{\extracolsep{\fill}}lcccccccc@{}}
\toprule
\textbf{Dataset} & \multicolumn{2}{c}{\textbf{HumanActivity}} & \multicolumn{2}{c}{\textbf{USHCN}} & \multicolumn{2}{c}{\textbf{PhysioNet}} & \multicolumn{2}{c}{\textbf{MIMIC}} \\
\midrule
\textbf{Metric} & \textbf{MSE} & \textbf{MAE} & \textbf{MSE} & \textbf{MAE} & \textbf{MSE} & \textbf{MAE} & \textbf{MSE} & \textbf{MAE} \\
\midrule
tPatchGNN & .0580$\pm$.0011 & .1448$\pm$.0027 & .5753$\pm$.0892 & .4111$\pm$.0713 & .3635$\pm$.0014 & .4120$\pm$.0020 & .5140$\pm$.0040 & .4440$\pm$.0052 \\
Hi-Patch & \textbf{.0557$\pm$.0002} & .1423$\pm$.0011 & .4528$\pm$.0075 & .3148$\pm$.0099 & .3628$\pm$.0018 & .4145$\pm$.0028 & .5024$\pm$.0115 & .4415$\pm$.0035 \\
KAFNet & .0559$\pm$.0002 & .1376$\pm$.0012 & .5327$\pm$.1160 & .3932$\pm$.0731 & .3709$\pm$.0018 & .4182$\pm$.0015 & .5334$\pm$.0253 & .4542$\pm$.0134 \\
APN & .0657$\pm$.0022 & .1577$\pm$.0033 & .5455$\pm$.1106 & .3894$\pm$.0642 & .3686$\pm$.0013 & .4159$\pm$.0020 & .4982$\pm$.0064 & .4421$\pm$.0070 \\
\midrule
\textbf{WrapFlow (Ours)} & .0591$\pm$.0011 & \textbf{.1373$\pm$.0004} & \textbf{.4086$\pm$.0111} & \textbf{.2571$\pm$.0063} & \textbf{.3569$\pm$.0012} & \textbf{.4013$\pm$.0045} & \textbf{.4847$\pm$.0015} & \textbf{.4049$\pm$.0013} \\
\bottomrule
\end{tabular*}
}
\caption{Experimental results on four irregular multivariate time series datasets evaluated using MSE and MAE, under extended forecast horizons while keeping the lookback settings consistent with the main-paper experiments: 12 hours for MIMIC-III and PhysioNet’12, 1000 milliseconds for HumanActivity, and 1 year for USHCN.}
\label{tab:varying_forecast_horizons}
\end{table*}

\begin{figure*}[t]
\centering
\includegraphics[width=\textwidth]{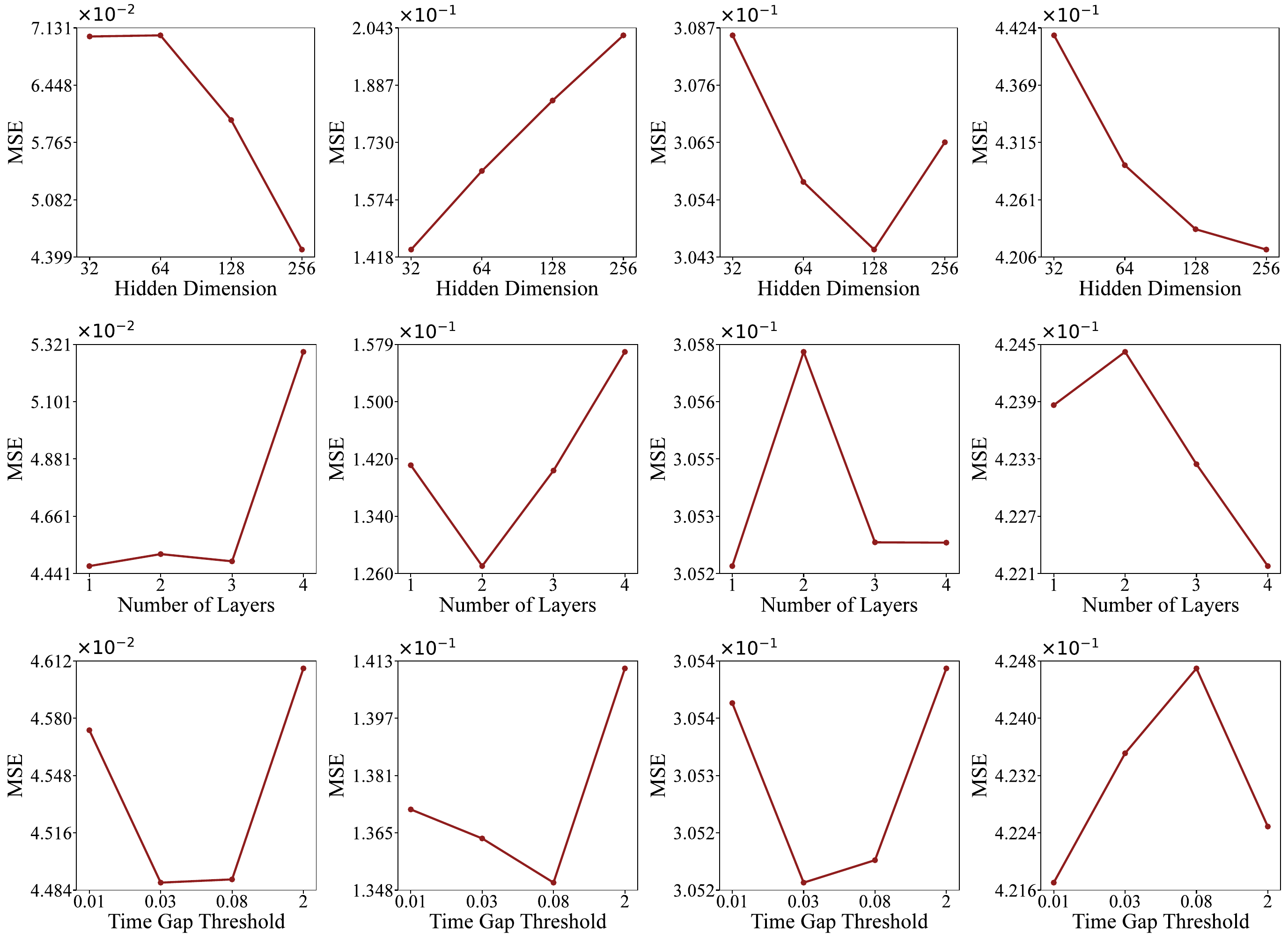}
\caption{Parameter sensitivity of WrapFlow on HumanActivity, USHCN, PhysioNet, and MIMIC. From top to bottom: hidden dimension $D$, number of encoder layers $L$, and time-gap threshold $\tau_{\mathrm{gap}}$. Each row reports MSE.}
\label{fig:app_param_sensitivity}
\end{figure*}

\subsection{Parameter Sensitivity}
\label{app:parameter_sensitivity}

We conduct parameter sensitivity studies for three key settings of WrapFlow: the hidden dimension $D$, the number of encoder layers $L$, and the time-gap threshold $\tau_{\mathrm{gap}}$.

\noindent\textbf{Hidden dimension.}
The first row of Figure~\ref{fig:app_param_sensitivity} shows that the effect of the hidden dimension $D$ reflects a trade-off between representation capacity and over-parameterization. A larger $D$ allows WrapFlow to preserve richer temporal patterns and cross-variable dependencies, which is beneficial for more complex datasets such as HumanActivity and MIMIC. In contrast, USHCN achieves the best performance with a smaller hidden dimension, suggesting that its temporal dynamics are relatively regular and do not require a large latent space. PhysioNet performs best at a moderate scale, indicating that an excessively large representation space is unnecessary. Overall, WrapFlow does not simply benefit from increasing model size; instead, the hidden dimension should match the intrinsic complexity of each dataset.

\noindent\textbf{Number of encoder layers.}
The second row shows that the effect of the number of encoder layers $L$ is dataset-dependent. HumanActivity and PhysioNet achieve the best performance with a single layer, while USHCN favors a slightly deeper setting with $L=2$. In contrast, MIMIC obtains the best result at $L=4$, indicating that deeper temporal abstraction can be beneficial for complex clinical observations. Overall, WrapFlow generally favors shallow-to-moderate depth, while the optimal encoder depth should be determined by the complexity of each dataset.

\noindent\textbf{Time-gap threshold.}
The third row shows that the time-gap threshold $\tau_{\mathrm{gap}}$ mainly controls how sensitively WrapFlow identifies informative observation gaps. If the threshold is too small, ordinary irregular intervals may be over-emphasized; if it is too large, truly meaningful long gaps may be ignored. The results show that WrapFlow is generally robust within a small-to-moderate range of thresholds, whereas excessively large values tend to degrade performance. This suggests that the GAP mechanism is most effective when it highlights informative silence without over- or under-detecting temporal gaps.

\end{document}